\documentclass[10pt,twocolumn,letterpaper]{article}

\usepackage{iccv}
\usepackage{times}
\usepackage{epsfig}
\usepackage{graphicx}
\usepackage{amsmath}
\usepackage{amssymb}

\usepackage{multirow,boldline}
\usepackage{siunitx}

\usepackage[pagebackref=true,breaklinks=true,letterpaper=true,colorlinks,bookmarks=false]{hyperref}

\iccvfinalcopy 

\def\iccvPaperID{5112} 

\ificcvfinal\fi
\begin{document}

\title{Towards Unconstrained End-to-End Text Spotting}
\author{Siyang Qin, Alessandro Bissacco, Michalis Raptis, Yasuhisa Fujii, Ying Xiao\\
Google AI\\
{\tt\small \{qinb,bissacco,mraptis,yasuhisaf,yingxiao\}@google.com}
}

\maketitle
\thispagestyle{empty}

\begin{abstract}
We propose an end-to-end trainable network that can simultaneously detect and recognize text of arbitrary shape, making substantial progress on the open problem of reading scene text of irregular shape. We formulate arbitrary shape text detection as an instance segmentation problem; an attention model is then used to decode the textual content of each irregularly shaped text region without rectification. To extract useful irregularly shaped text instance features from image scale features, we propose a simple yet effective RoI masking step. Additionally, we show that predictions from an existing multi-step OCR engine can be leveraged as partially labeled training data, which leads to significant improvements in both the detection and recognition accuracy of our model. Our method surpasses the state-of-the-art for end-to-end recognition tasks on the ICDAR15 (straight) benchmark by 4.6\%, and on the Total-Text (curved) benchmark by more than 16\%.
\end{abstract}

\section{Introduction}\label{sec:intro}
Automatically detecting and recognizing text in images can benefit a large number of practical applications, such as autonomous driving, surveillance, or visual search and can increase the environmental awareness of visually impaired people \cite{neat2019scene}.

Traditional optical character recognition (OCR) pipeline methods usually partition the scene text reading task into two sub-problems, \emph{scene text detection} and \emph{cropped text line recognition}. Text detection methods try to spot text instances (words or lines) in the input image, while text recognition models take a cropped text patch and decode its textual content. Since most scene text detection methods are unable to directly predict the correct text reading direction, an additional direction identification step is necessary for successful OCR engines \cite{walker2018web}. 

Despite their long history and great success, the use of multiple models within an OCR pipeline engine has several disadvantages: errors can accumulate in such a cascade which may lead to a large fraction of garbage predictions. Furthermore, each model in the pipeline depends on the outputs of the previous step, which makes it hard to jointly maximize the end-to-end performance, and fine-tune the engine with new data or adapt it to a new domain. Finally, maintaining such a cascaded pipeline with data and model dependencies requires substantial engineering effort.

\begin{figure}[t]
\begin{center}
   \includegraphics[width=1.0\linewidth]{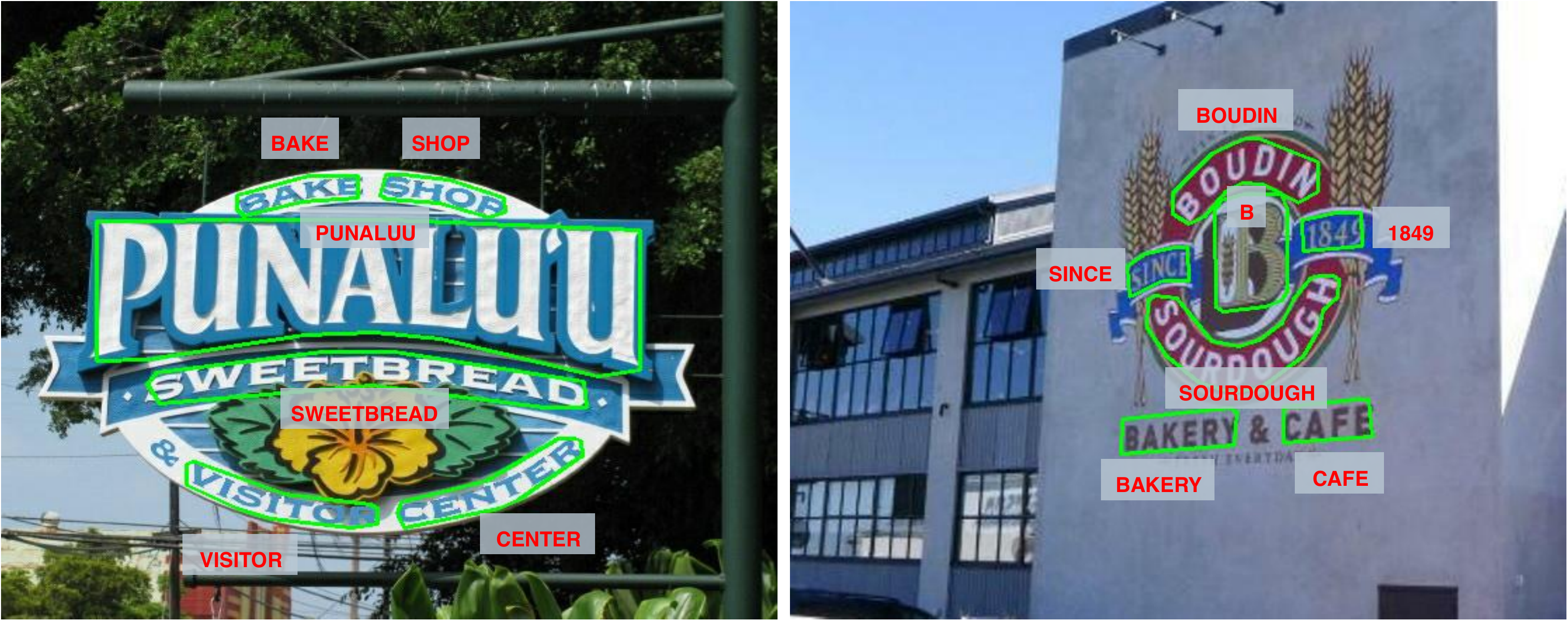}
\end{center}
   \caption{Our end-to-end model can predict the locations and transcriptions of text with arbitrary shape in a single forward pass.}
\label{fig:teaser}
\end{figure}

End-to-end OCR models overcome those disadvantages and thus have recently started gaining traction in the research community \cite{lyu2018mask,sun2018textnet,liu2018fots,he2018end,li2017towards}. The basic idea behind end-to-end OCR is to have the detector and recognizer share the same CNN feature extractor. During training, the detector and recognizer are jointly optimized; at inference time, the model can predict locations and transcriptions in a single forward pass. While producing superior accuracy in straight text reading benchmarks, these methods struggle to generalize and produce convincing results on more challenging datasets with curved text, which arise naturally and frequently in everyday environments (see Figure \ref{fig:teaser} for two such examples). Handling arbitrary shaped text is a crucial open problem in order for OCR to move beyond its traditional straight text applications.

In this paper, we propose a simple and flexible end-to-end OCR model based on a Mask R-CNN detector and a sequence-to-sequence (seq2seq)  attention decoder \cite{bahdanau2014neural}. We make no assumptions on the shape of the text: our model can detect and recognize text of arbitrary shape, not just the limited case of straight lines. The key idea underlying our model is to skip the feature rectification step between the detector and the recognizer, 
and directly feed cropped and masked text instance features to the decoder. We show that our model is able to recognize text in different orientations and even along curved paths. Our model learns where to start decoding, and how to update the attention weights to follow the unrectified text path. Our detector is based on Mask R-CNN: for each text instance, it predicts an axis-aligned rectangular bounding box and the corresponding segmentation mask. Using these, our model works seamlessly on both straight and curved text paths.

Typically the recognizer requires far more data to train than the detector. Unlike the case of multi-step OCR models where cropped text lines (easier to collect and synthesize) are used to train the recognizer, previous end-to-end models demand fully labeled images as training data. This makes end-to-end training challenging due to the short of fully annotated images. Furthermore, by the time the recognizer has converged, the detector is often substantially overfitted. In this work, we solve both issues by adding additional large scale partially labeled data which is automatically labeled by an existing multi-step OCR engine \footnote{Publicly available via Google Cloud Vision API.} \cite{bissacco2013photoocr}. If an input training sample is partially annotated, only the recognizer branch is trained. We find that this significantly boosts the performance of our model.

Our method surpasses the previous state-of-the-art results by a large margin on both straight and curved OCR benchmarks. On the popular and challenging ICDAR15 (straight) dataset, our model out-performs the previous highest by 4.6\% on end-to-end F-score. On the Total-Text (curved) dataset, we significantly increase the state-of-the-art by more than 16\%.

In summary, the contributions of this paper are three-fold:
\begin{itemize}
    \item We propose a flexible and powerful end-to-end OCR model which is based on Mask R-CNN and attention decoder. Without bells and whistles, our model achieves state-of-the-art results on both straight and curved OCR benchmarks.
    \item We identify feature rectification as a key bottleneck in generalizing to irregular shaped text, and introduce a simple technique (RoI masking) that makes rectification unnecessary for the recognizer. This allows the attention decoder to directly operate on arbitrarily shaped text instances.
    \item To the best of our knowledge, this is the first work to show that end-to-end training can benefit from partially labeled data bootstrapped from an existing multi-step OCR engine.
\end{itemize}

\section{Related Work}\label{sec:relatedwork}
In this section, we briefly review the existing text detection and recognition methods, and highlight the differences between our method and current end-to-end models. For a more detailed review, the reader is referred to \cite{long2018scene}.

\noindent \textbf{Scene Text Detection:}
Over the years, the traditional sliding window based methods \cite{jaderberg2014deep,chen2004detecting,zhu2016text} and connected-component based methods \cite{busta2015fastext,huang2013text,neumann2013scene,neumann2012real,epshtein2010detecting,qin2016fast} have been replaced by deep learning inspired methods with a simplified pipeline. These newer methods have absorbed the advances from general object detection\cite{liu2016ssd,ren2015faster,redmon2016you} and semantic segmentation \cite{long2015fully,chen2018deeplab} algorithms, adding well-designed modifications specific to text detection. Modern scene text detection algorithms can directly predict oriented rectangular bounding boxes or tighter quadrilaterals via either single-shot \cite{liao2017textboxes,zhou2017east,he2017single,liu2017deep}, or two-stage models \cite{jiang2017r2cnn,liao2018rotation,ma2018arbitrary,qin2017cascaded}.

Recently, detecting curved text in images has become an emerging topic: a new dataset containing curved text was introduced in \cite{ch2017total} which provides tight polygon bounding boxes and ground-truth transcriptions. In \cite{dai2018fused}, Dai \etal formulate text detection as an instance segmentation problem, and in \cite{long2018textsnake}, the authors proposed representing a text instance as a sequence of ordered, overlapping disks, which is able to cover curved cases. Despite the success in detecting curved text, \emph{reading} the curved text is an unsolved problem.

\noindent \textbf{Scene Text Recognition:}
The goal of scene text recognition algorithms is to decode the textual content from \emph{cropped text patches}. Modern scene text recognition methods can be grouped into two main categories, CTC (Connectionist Temporal Classification \cite{graves2006connectionist}) based methods \cite{shi2017end,he2016reading,fujii2017sequence} and attention based methods \cite{wojna2017attention,cheng2017focusing,cheng2018aon,ghosh2017visual,lee2016recursive,shi2016robust}. Most scene text recognition methods from both categories assume the input text is rectified (straight line, read from left to right): the input is first resized to have a constant height, then fed to a fully convolutional network to extract features. To capture long range sequence context, some CTC-based architectures stack an RNN on top of a CNN, while others use stacked convolution layers, with a large receptive field. Finally, each feature column predicts a symbol and duplicated symbols are removed to produce the final prediction. In attention models, RNN is often used to predict one symbol per step based on the prediction at the previous step, the hidden state, and a weighted sum of the extracted image features (context). The process stops when the end-of-sequence symbol is predicted, or the maximum number of iterations is reached.

\begin{figure*}
\begin{center}
    \includegraphics[width=1.0\linewidth]{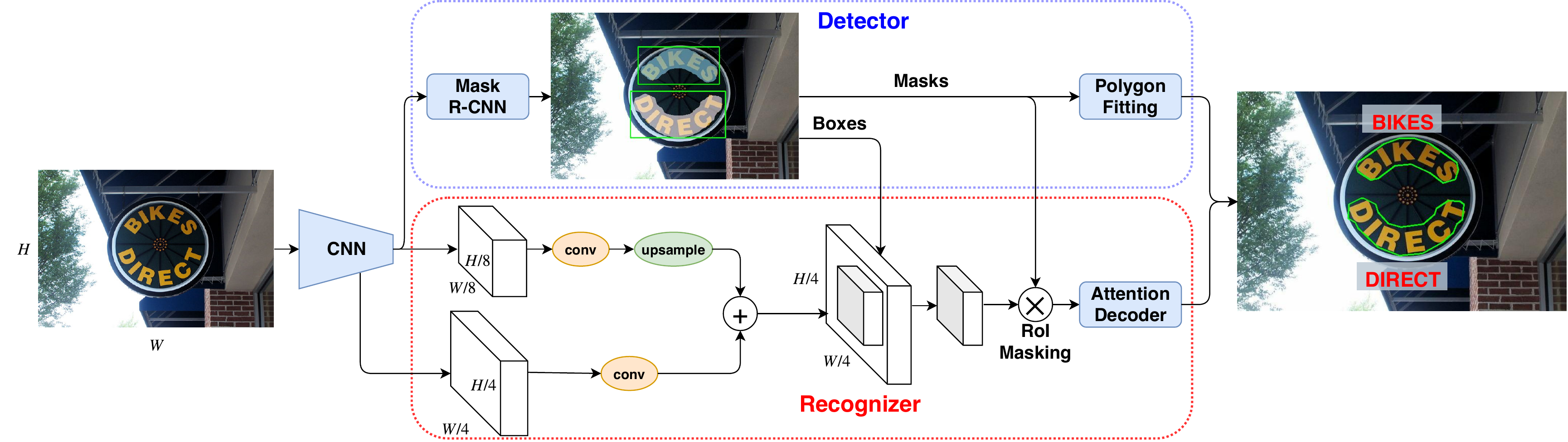}
\end{center}
   \caption{Overall architecture of our end-to-end OCR model.}
\label{fig:pipeline}
\end{figure*}

\noindent \textbf{End-to-End OCR:}
The work from Li \etal \cite{li2017towards} is the first successful end-to-end OCR model which only supports horizontal text. 
The multi-oriented end-to-end OCR architectures of Liu \etal \cite{liu2018fots}, He \etal \cite{he2018end} and Sun \etal \cite{sun2018textnet} share a common idea: they feed \emph{rectified text region features} to the recognizer to enable end-to-end training. In \cite{liu2018fots}, the model outputs rotated rectangles in the detection stage, and used a CTC-based recognizer which can not generalize to curved cases. In \cite{sun2018textnet}, the detector outputs quadrilaterals and an attention-based model is used to decode the textual content. In contrast, our detector produces rectangular bounding boxes and the corresponding instance segmentation masks, which is a more general way to represent text in arbitrary shape. In addition, we remove the feature rectification step which is designed for straight text, and let the attention decoder directly operates on cropped and masked text instance features. This leads to better flexibility and performance in curved text.

Lyu \etal \cite{lyu2018mask} proposed an end-to-end OCR engine which is based on Mask R-CNN. They adopted a simple recognition by detection scheme: in order to recognize the text, all the characters are detected individually. This method is not ideal because much of the sequential information is lost. Furthermore, detecting individual character can be difficult or even impossible in many cases. And even if all the characters are correctly detected, it is highly unclear how to link them into a correct sequence. In \cite{lyu2018mask}, the authors simply group characters from left to right, which precludes correct recognition of text in non-traditional reading directions. On the other hand, by leveraging sequential information, our method is able to correctly recognize text in more challenging situations and non-traditional reading directions.

\section{Model Architecture and Training}\label{sec:method}

Figures~\ref{fig:pipeline} shows the design of our end-to-end OCR model. The detector part of the model is based on Mask R-CNN which has been widely used in instance segmentation and other related tasks. For each text region (word or text line), Mask R-CNN can predict an axis-aligned rectangular bounding box and the corresponding instance segmentation mask. For the straight text case, the final detection results are obtained by fitting a min-area rotated rectangle to each segmentation mask, while a general polygon is fitted to each mask for the curved text case. By using Mask R-CNN as the detector, our model works seamlessly with straight and curved text paths.

A novel feature of our architecture is that \emph{we do not rectify the input to the recognizer}. This renders traditional CTC-based decoders unsuitable. Instead, we use a seq2seq model (with attention) as recognizer. 
At each step, the decoder makes the prediction based on the output and state from the previous step, as well as a convex combination of the text instance features (context). In order to extract arbitrary shaped text instance features from image level features, we introduce \emph{RoI masking} which multiplies the cropped features with text instance segmentation masks. This removes neighboring text and background, and ensures that the attention decoder will focus only on the current text instance.

\subsection{Feature Extractor} \label{sec:cnn}
We explore two popular backbone architecture, ResNet-50\cite{he2016deep} and Inception-ResNet \cite{szegedy2017inception}; the latter model is far larger, and consequently yields better detection and recognition results. Scene text usually has large variance in scale; in order to capture both large and tiny text, the backbone should provide dense features while maintaining a large receptive field. To achieve this, we follow the suggestions from \cite{huang2017speed}: both backbones are modified to have an effective output stride of 8. In order to maintain a large receptive field, atrous convolutions are used to compensate for the reduced stride.

For ResNet-50, we modified the \textit{conv4\_1}\footnote{Our naming convention follows \cite{huang2017speed}.} layer to have stride 1 and use atrous convolution for all subsequent layers. We extract features from the output of the third stage. The Inception-ResNet is modified in a similar way to have output stride 8, taking the output from the second repeated block (layer \textit{PreAuxLogits}).

\subsection{Detector} \label{sec:detector}
We follow the standard Mask R-CNN implementation. In the first stage, a region proposal network (RPN) is used to propose a number of candidate text regions of interest (RoIs). In the second stage, each RoI is processed by three prediction heads: a class prediction head to decide if it is text or not, a bounding box regression head to predict an axis-aligned rectangular box, and finally a mask prediction head to predict the corresponding instance segmentation mask.

The RPN anchors span four scales (64, 128, 256, 512) and three aspect ratios (0.5, 1.0, 2.0); using more scales and aspect ratios may increase the model's performance at the cost of longer inference time. Non-maximum suppression (NMS) is used to remove highly overlapping proposals with intersection-over-union (IoU) threshold set to 0.7. The top 300 proposals are kept.  In the second stage, features from each RoI are cropped and resized to $28\times28$ followed by a $2\times2$ max pooling, which lead to $14\times14$ features for each RoI. At training time, RoIs are grouped into mini batches of size 64, and then fed to a class prediction head and bounding box refinement head. A second NMS is performed on top of refined boxes (IoU is set to 0.7). During inference time, the top 100 regions are sent to the mask prediction head. The final detection output is obtained after the final NMS step, which computes the IoU based on the mask instead of bounding boxes like the first two NMS steps.

\subsection{Multi-Scale Feature Fusion and RoI Masking} \label{sec:roimask}
In our experiments, we found that stride 8 features and multi-scale anchors are sufficient for the text detection task for both large and small text. However, for text recognition, a finer-grained task, denser features are needed. Inspired by the feature pyramid network \cite{lin2017feature}, we gradually upsample lower resolution, but context rich features, and fuse them with higher resolution features from earlier CNN layers using element-wise addition. A $1\times 1$ convolution (with 128 channels) is applied to all features to reduce dimensionality, and to ensure uniform shapes,  before element-wise addition. This produces a dense feature map which encodes both local features and longer range contextual information, which can improve recognition performance especially for small text. In practice, we find that fusing features with stride 8 and 4 leads to the best results. More specifically, for ResNet-50, we use features after the first (stride 4), second (stride 8) and third stage (stride 8); the corresponding receptive field sizes are 35, 99 and 291 respectively. For Inception-ResNet, we use features after layer \textit{Conv2d\_4a\_3x3} (stride 4), \textit{Mixed\_5b} (stride 8) and \textit{PreAuxLogits} (stride 8), the corresponding receptive field sizes are 23, 63 and 2335 respectively.

In multi-step OCR engines, each text instance is cropped out from the input image before being fed to the recognizer. In contrast, in end-to-end models, instead of cropping out the image patch, a more involved method is used to extract text instance features from the image level features output by the backbone CNN. For object detection models, axis-aligned bounding boxes are used to crop out features \cite{girshick2015fast}. For text, the work in \cite{liu2018fots} and \cite{sun2018textnet} proposed RoI rotate and perspective RoI transforms to compute rectified text instance features using rotated rectangles or quadrilaterals. This works well for straight text but fails in the curved text case. In this work, we propose a simple and more general way to extract text instance features that works for any shape, called \emph{RoI masking}: first the predicted axis-aligned rectangular bounding boxes are used to crop out features, and then we multiply by the corresponding instance segmentation mask. Since we \emph{do not} know the reading direction of the text at this point, features from each region are resized so that the shorter dimension is equal to 14 while maintaining the overall aspect ratio. RoI masking filters out the neighboring text and background, ensuring that the attention decoder will not accidentally focus on areas outside the current decoding region. Our ablation experiments in Section \ref{sec:ablation} show that RoI masking substantially improves the recognizer's performance.

\begin{figure}[t]
\begin{center}
   \includegraphics[width=1.0\linewidth]{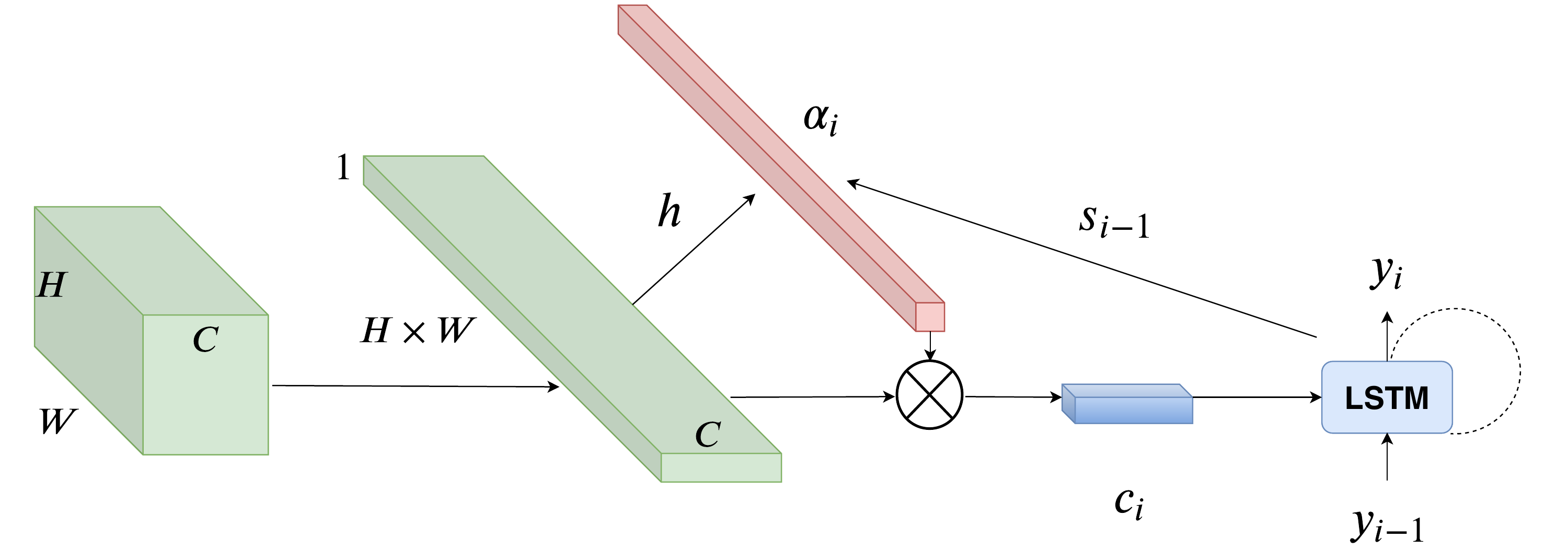}
\end{center}
   \caption{Our seq2seq based recognizer.}
\label{fig:recognizer}
\end{figure}

\subsection{Recognizer} \label{sec:recognizer}

The recognizer is a seq2seq model with Bahdanau-style attention proposed in \cite{bahdanau2014neural}, shown in Figure \ref{fig:recognizer}. At the first step, the model takes a \textit{START} symbol and zero LSTM initial state; we then produce symbols until the End-of-Sequence (\textit{EOS}) symbol is predicted. At each step, the final predicted distribution over possible symbols is given by:

\begin{equation}
    p(y_i|y_1,...,y_{i-1},h)=\textrm{softmax}(W_oo_i+b_o)   
\end{equation}

Where $y_i$ is the predicted character, $o_i$ is the LSTM output at time step $i$ respectively, and $h$ represents the flattened extracted text instance features. At each step, the LSTM takes the prediction of the previous step $y_{i-1}$, the previous hidden state $s_{i-1}$ and a weighted sum of the image feature $c_i$ (context) to compute the output $o_i$ and new state vector $s_i$.

\begin{equation}
    (o_i, s_i)=LSTM(y_{i-1}, s_{i-1}, c_i)
\end{equation}

At each step, the decoder is able to pay attention to some specific image region and use the corresponding image features to help make the right prediction. The context vector $c_i$ is a weighted sum of the flattened image feature $h$ and learned weight vector $\alpha_i$: $c_i=\sum_j \alpha_{ij}h_j$. The weight vector $\alpha_i$ is defined as:

\begin{equation}
    \alpha_{ij}=\frac{\exp(e_{ij})}{\sum_k \exp(e_{ik})}
\end{equation}

\begin{equation}
    e_{ij}=V^T\tanh(W_s s_{i-1}+W_h h_j).
\end{equation}

The attention weight for each feature position is determined by image feature ($h$) and previous LSTM state ($s_{i-1}$) which encode the shift of the attention mask. This enables the recognizer to follow arbitrary shaped text lines.

By feeding the predicted symbol to the next step, the model can learn an implicit language model. At inference time, the predicted symbol is fed to the next step while the ground-truth one is used during training (i.e., teacher forcing).

\subsection{Joint Training and Loss Function} \label{sec:loss}
We observe that the recognizer requires far more data and training iterations in comparison to the detector; this makes joint training difficult as the existing public datasets are not large enough to train a high performance attention decoder, especially when the input features are not rectified. Furthermore, if we train long enough to achieve convergence in the recognizer, there is a strong risk of overfitting the detector. In this work, we solve both issues by adding additional large scale partially labeled data which is automatically labeled by an existing multi-stage OCR engine from Google Cloud Vision API. If the input training sample is fully labeled, we update the weights of both detector and recognizer. If it has been automatically annotated by an OCR engine (and thus may have unlabeled text), only the recognizer branch is trained. Thus the total multitask loss is defined as:

\begin{equation}
    L=\delta(L_{rpn}+\alpha L_{rcnn}+\beta L_{mask})+\gamma L_{recog}.
\end{equation}

Here, $\delta$ is $1$ if the input is fully labeled, $0$ otherwise. In our implementation, both $\alpha$, $\beta$ and $\gamma$ are set to $1.0$. Adding machine labeled, partially labeled data ensures that the recognizer ``sees'' enough text while preventing the detector from overfitting. For the machine labeled data, since the detection of all the text is not required, we could increase the confidence threshold to filter out noisy low confidence outputs.

The detection losses are the same as the original Mask R-CNN paper \cite{he2017mask}. The recognizer loss $L_{recog}$ is the cross entropy loss with label smoothing set to $0.9$, as suggested by \cite{wojna2017attention}. During training, the ground-truth boxes and masks are used for RoI cropping and RoI masking while the predicted ones are used at inference time. We also tried to use predicted bounding boxes and masks during training but found no improvement.

\subsection{Implementation Details} \label{sec:details}
The data used to train our model contains images from the training portion of popular public datasets, including SynthText, ICDAR15, COCO-Text, ICDAR-MLT and Total-Text. The number of images we used from each dataset are 200k, 1k, 17k, 7k and 1255 respectively. Besides public datasets, we also collected 30k images from the web and manually labeled each word, providing oriented rectangular bounding boxes and transcriptions. The number of fully labeled real images is too low to train a robust end-to-end OCR model. To solve this issue, as mentioned in Section \ref{sec:loss}, we run an existing OCR engine on one million images with text and use the predictions (oriented rectangles and transcriptions) as the partially labeled ground-truth. Our experiments (see Section \ref{sec:ablation}) show this can significantly improve the end-to-end performance. To prevent the large volumes of synthetic and partially labeled data from dominating the training data, extensive data augmentations are applied to fully labeled real images. First, the shorter dimension of input image is resized from 480 to 800 pixels, then random rotation, random cropping and aspect ratio jittering are used.

We adopt a single-step training strategy. The backbone CNN is pre-trained on ImageNet; the detector and recognizer are jointly optimized using both fully and partially annotated images. Our ablation experiment (see Section \ref{sec:ablation}) shows that this achieves better accuracy than a two-step training strategy, where we first use all the fully labeled data to train the detector and then jointly fine-tune the detector and recognizer using both fully and partially labeled data. We train our model with asynchronous SGD with momentum of 0.9. The initial learning rate depends on the backbone network, $10^{-3}$ for Inception-ResNet and $3\times10^{-4}$ for ResNet-50. We reduce the learning rate by a factor of 3 every 2M iterations, with a total number of 8M iterations. During training, each GPU takes a single training sample per iteration, and 15 Tesla V100 GPUs are used. We implement the model using TensorFlow \cite{abadi2016tensorflow}, the training process takes about three days to finish. In the recognizer, we use a single layer LSTM with 256 hidden units. Recurrent dropout \cite{gal2016theoretically} and layer normalization \cite{ba2016layer} are used to reduce overfitting. The total number of symbols are 79, which includes digits, upper and lower cases of English characters as well as several special characters.

\begin{table*}[t]
\begin{center}
\resizebox{0.9\textwidth}{!}{
\begin{tabular}{c|ccc|c|ccc}

\hlineB{2.0}
\multirow{2}{*}{Method} & \multicolumn{3}{c|}{Detection} & \multirow{2}{*}{Method} & \multicolumn{3}{c}{End-to-End} \\ \cline{2-4} \cline{6-8} 
                                & P & R & F &  & S  & W & G \\ \hlineB{2.0}
SSTD \cite{he2017single}        & 80.23 & 73.86 & 76.91 & Stradvision \cite{karatzas2015icdar} & 43.70 & - & - \\ 
EAST \cite{zhou2017east}        & 83.27 & 78.33 & 80.72 & TextProposals+DictNet \cite{gomez2017textproposals,jaderberg2014synthetic} & 56.0 & 52.3 & 49.7 \\ 
TextSnake \cite{long2018textsnake} & 84.9 & 80.4 & 82.6 & HUST\_MCLAB \cite{shi2017detecting,shi2017end} & 67.86 & - & - \\ 
RRD MS \cite{liao2018rotation}  & 88 & 80 & 83.8 & E2E-MLT \cite{buvsta2018e2e} & - & - & 55.1 \\ 
Mask TextSpotter \cite{lyu2018mask} & 91.6 & 81.0 & 86.0 & Mask TextSpotter \cite{lyu2018mask} & 79.3 & 73.0 & 62.4 \\ 
TextNet \cite{sun2018textnet}   & 89.42 & 85.41 & 87.37 & TextNet \cite{sun2018textnet} & 78.66 & 74.90 & 60.45 \\ 
He \etal \cite{he2018end}   & 87 & 86 & 87 & He \etal \cite{he2018end} & 82 & 77 & 63 \\ 
FOTS \cite{liu2018fots}         & 91.0 & 85.17 & 87.99 & FOTS \cite{liu2018fots} & 81.09 & 75.90 & 60.80 \\ 
FOTS MS \cite{liu2018fots}      & \textbf{91.85} & 87.92 & \textbf{89.84} & FOTS MS \cite{liu2018fots} & 83.55 & 79.11 & 65.33 \\ \hline
Ours (ResNet-50) & 89.36 & 85.75 & 87.52 & Ours (ResNet-50) & 83.38 & 79.94 & 67.98 \\  
Ours (Inception-ResNet) & 91.67 & \textbf{87.96} & 89.78 & Ours  (Inception-ResNet)      & \textbf{85.51} & \textbf{81.91} & \textbf{69.94} \\ \hlineB{2.0}

\end{tabular}}
\end{center}
\caption{Comparison on ICDAR15. ``MS'' represents multi-scale testing. ``P'', ``R'' and ``F'' stand for precision, recall, and F-score respectively. In the end-to-end evaluation, F-score under three lexicon settings are shown. ``S'' (strong) means 100 words, including the ground-truth, are given for each image. For ``W'' (weak), a lexicon includes all the words appeared in the test set is provided. For ``G'', a generic lexicon with 90k words is given, which is not used by our model.}
\label{table:IC15}
\end{table*}

\section{Experiments}
We evaluate the performance of our model on the ICDAR15 benchmark \cite{karatzas2015icdar} (straight text) and recently introduced Total-Text \cite{ch2017total} (curved text) dataset.

\subsection{Straight Text}

\begin{figure*}
\begin{center}
    \includegraphics[width=1.0\linewidth]{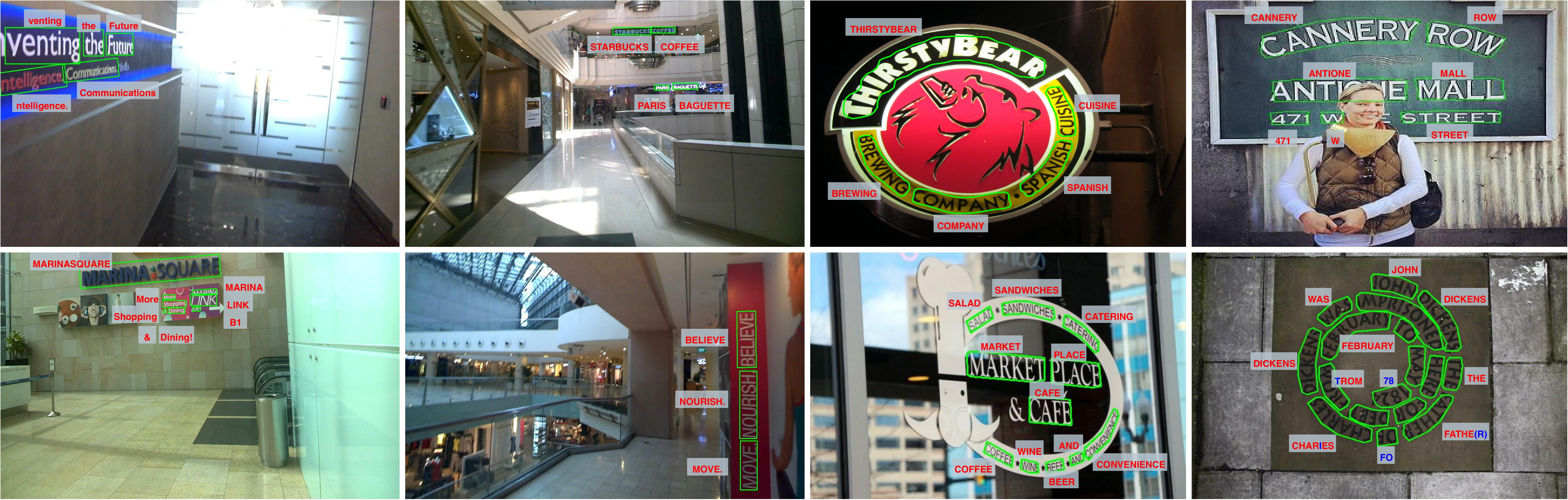}
\end{center}
   \caption{Qualitative results of our method on ICDAR15 (first two columns) and Total-Text (last two columns) datasets. In the bottom right image, prediction errors are shown in blue, some predictions are skipped for better visualization. All the skipped predictions are correctly predicted by our method.}
\label{fig:samples}
\end{figure*}

We show the superior performance of our model on detecting and recognizing oriented straight text using the ICDAR15 benchmark introduced in Challenge 4 of the ICDAR 2015 Robust Reading Competition. The dataset consists of 1000 training images and 500 testing images. Images in this dataset are captured by wearable cameras, without intentional focus on text regions. There are large variations in text size, orientation, font, and lighting conditions. Motion blur is also common. In this dataset, text instances are labeled at the word level. Quadrilateral bounding boxes and transcriptions are provided. For detection evaluation, a prediction is counted as a true positive if the IoU with the closest ground-truth is larger than 0.5. For end-to-end evaluation, the predicted transcription needs to be identical to the corresponding ground-truth in order to be considered as a true positive. Some unreadable words are marked as ``do not care''. The Evaluation metrics of interest are precision (true positives count over detection count), recall (true positives count over ground-truth count), and F-score (harmonic mean of precision and recall).

The results are summarized in Table \ref{table:IC15}. At inference time, the shorter dimension of the image is resized to 900 pixels. Note we only use a single scale input. In the detection only task, our method (with Inception-ResNet backbone) surpasses the best single scale model (FOTS) by 1.8\%. For end-to-end performance, our method out-performs the highest single scale model (He \etal) by about 7\%. Compared to the multi-scale version of the FOTS model, the current state-of-the-art, our method matches the detection performance while still achieving 4.6\% higher end-to-end F-score.

\subsection{Curved Text}
The biggest advantage of our method is the outstanding performance on irregular shaped text. We conducted an experiment on the recently introduced curved text dataset called Total-Text \cite{ch2017total}. Total-Text contains 1255 images for training and another 300 for testing, with a large number of curved text. In each image, text is annotated at word level, each word is labeled by a bounding polygon. Ground truth transcriptions are provided. The evaluation protocol for detection is based on \cite{wolf2006object}, the one for end-to-end recognition is based on ICDAR15's end-to-end evaluation protocol (the evaluation script is modified to support general polygons).

\begin{table}
\begin{center}
\resizebox{0.45\textwidth}{!}{
\begin{tabular}{c|ccc|c}
\hlineB{2.0}
\multirow{2}{*}{Method} & \multicolumn{3}{c|}{Detection} & E2E \\ \cline{2-5} 
                        & P        & R        & F        & None          \\ \hlineB{2.0}
Baseline \cite{ch2017total} & 40.0 & 33.0 & 36.0 & -  \\ 
Textboxes \cite{liao2017textboxes} & 62.1 & 45.5 & 52.5 & 36.3 \\ 
TextSnake \cite{long2018textsnake} & 82.7 & 74.5 & 78.4 & - \\ 
MSR \cite{xue2019msr} & 85.2 & 73.0 & 78.6 & - \\ 
TextField \cite{xu2019textfield} & 81.2 & 79.9 & 80.6 & - \\ 
FTSN \cite{dai2018fused} & 84.7 & 78.0 & 81.3 & - \\ 
Mask TextSpotter \cite{lyu2018mask} & 69.0 & 55.0 & 61.3 & 52.9 \\ 
TextNet \cite{sun2018textnet} & 68.21 & 59.45 & 63.53 & 54.0 \\ \hline
Ours (ResNet-50) & 83.3 & 83.4 & 83.3 & 67.8 \\ 
Ours (Inc-Res public) & 86.8 & 84.3 & 85.5 & 63.9 \\ 
Ours (Inc-Res) & \textbf{87.8} & \textbf{85.0} & \textbf{86.4} & \textbf{70.7} \\ \hlineB{2.0}
\end{tabular}}
\end{center}
\caption{Results on Total-Text. No lexicon is used in end-to-end evaluation. ``Inc-Res'' stands for Inception-Resnet. ``Inc-Res public'' represents our model with Inception-ResNet backbone, trained using only public datasets.}
\label{table:Total-Text}
\end{table}

During training, we pre-train the model on straight text and fine-tune it using \emph{only images from the training portion of the Total-Text dataset}. At inference time, the shorter dimension of each image is resized to 600 pixels. We compare the results of our model with both backbones against previous work in Table \ref{table:Total-Text}. We also list results \emph{trained using only publicly available datasets} for the Inception-ResNet backbone. Our method out-performs the previous state-of-the-art by a large margin in both detection and end-to-end evaluations. Specifically, for detection, our best model surpasses the previous highest by 5.1\%. In the end-to-end recognition task, our best model significantly raises the bar by 16.7\%. In the absence of our internal fully labeled data and partially machine annotated data, our method still achieves far better performance in both the detection and recognition tasks, at +4.2\% and +9.9\% respectively.


Several qualitative examples are shown in Figure \ref{fig:samples} (third and fourth column). Our method produces high quality bounding polygons and transcriptions. Surprisingly, we find that our method can also produce reasonable predictions in partially occluded cases (top right image, ``ANTIONE'') by utilizing visible image features and the learned implicit language model from the LSTM. In the bottom right image, we show some failure cases, where the text is upside down and read from right to left. These cases are very rare in the training data, we believe more aggressive data augmentation may mitigate these issues.

\begin{figure*}
\begin{center}
    \includegraphics[width=1.0\linewidth]{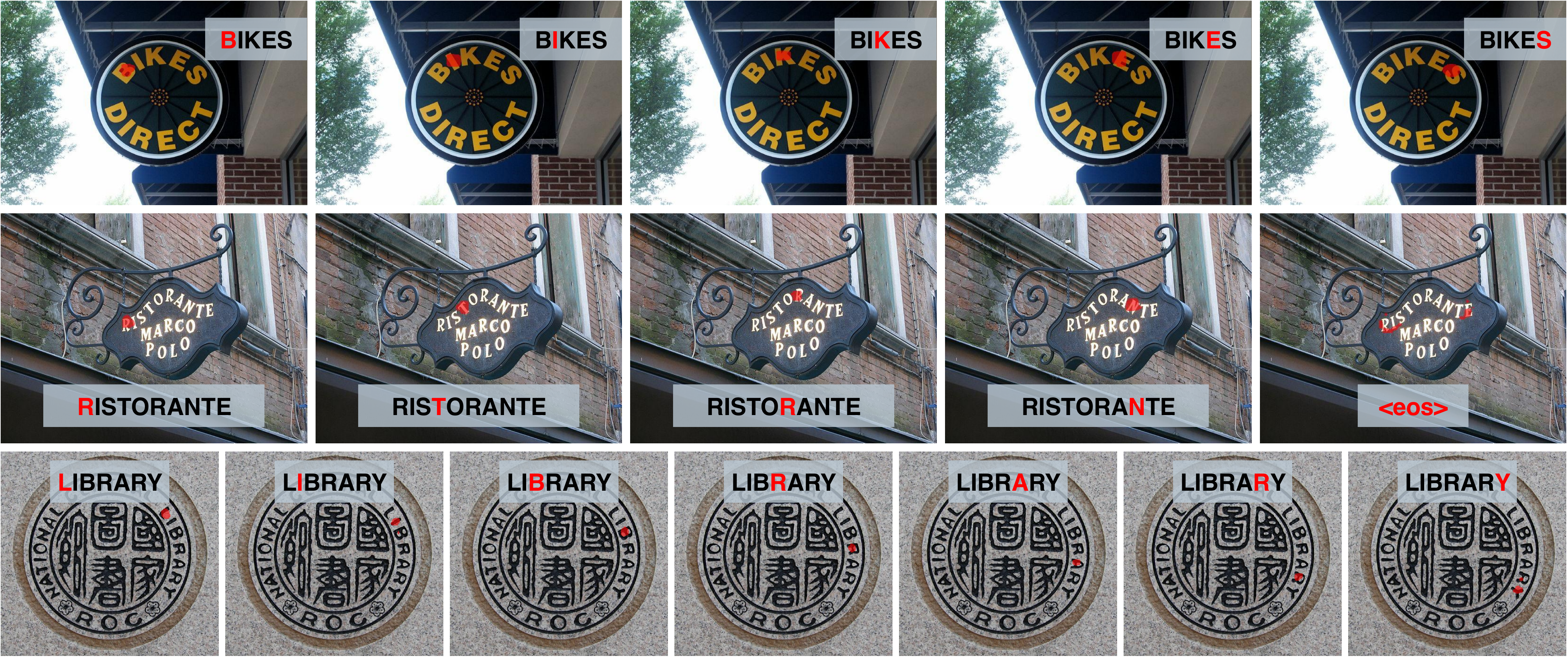}
\end{center}
   \caption{Visualization of the attention weights. Some steps are skipped for better visualization.}
\label{fig:attn}
\end{figure*}

We can visualize the attention weight vector at each step by reshaping it to 2D and projecting back to image coordinates. This provides a great tool to analyze and debug the model performance. In Figure \ref{fig:attn}, we find that the seq2seq model focuses on the right area when decoding each symbol and is able to follow the shape of the text. In the middle row (last image), we show the attention mask corresponding to the \textit{EOS} symbol. The attention is spread across both the start and end positions. 

\subsection{Ablation Experiments} \label{sec:ablation}
We conduct a series of ablation experiments to better understand our end-to-end OCR model. In this section, we report \emph{average precision} (AP) score, which is often a better evaluation metric than F-score (which display sensitivity to a specific threshold).
We use the ICDAR15 test set in this section. Table~\ref{table:onevstwo} summarizes the experimental results.

\noindent \textbf{Baselines:} We build a \emph{detection-only baseline} (first row in Table~\ref{table:onevstwo}) and an \emph{end-to-end baseline} (third row in Table~\ref{table:onevstwo}). In the detection-only baseline, we train a model with only the detection branch. In the end-to-end baseline, we train a model with both detection and recognition branches, but do not use partially labeled data or RoI masking, and adopt a single-step training strategy (described in Section ~\ref{sec:details}). The end-to-end baseline exhibits stronger detection results than the detection-only baseline (with a ResNet-50 backbone, the improvement is 1.6\%) despite being trained on exactly the same data. This suggests that training a recognizer improves the feature extractor for the detection task.

\noindent \textbf{Backbones:} From Table~\ref{table:onevstwo} we find that on the detection task, the more powerful Inception-ResNet backbone consistently out-performs ResNet-50. On the end-to-end task, our model with the ResNet-50 backbone actually achieves better performance when the training data is limited (without large scale partially labeled data). For our \emph{full end-to-end model}, the Inception-ResNet backbone achieves marginal improvement on end-to-end AP score (59.5\% \vs 59.0\%).

\begin{table}[t]
\begin{center}
\resizebox{0.48\textwidth}{!}{
\begin{tabular}{c|c|cc|cc|cc}
\hlineB{2.0}
\multicolumn{2}{c|}{\multirow{2}{*}{}}         & \multirow{2}{*}{PD} & \multirow{2}{*}{Mask} & \multicolumn{2}{c|}{ResNet-50} & \multicolumn{2}{c}{Inc-Res} \\ \cline{5-8} 
\multicolumn{2}{c|}{} &  &  & $AP_{Det}$ & $AP_{E2E}$ & $AP_{Det}$ & $AP_{E2E}$       \\ \hlineB{2.0}
\multirow{2}{*}{Two-step} & Det-baseline                  &  &  & 85.5 & - & 88.2 & -          \\ 
& E2E-full                  & \checkmark & \checkmark & 86.9 & 55.3 & 89.1 & 57.4 \\ \hline
\multirow{4}{*}{Single-step} & E2E-baseline & &  & 87.1 & 52.8 & 88.2 & 51.7    \\
 & + Mask &  & \checkmark & 86.7 & 53.9 & 88.9 & 53.1           \\
 & + PD & \checkmark & & \textbf{87.5} & 55.7 & 89.9 & 58.7           \\
 & E2E-full & \checkmark & \checkmark & 87.2 & \textbf{59.0} & \textbf{90.8} & \textbf{59.5} \\ \hlineB{2.0}
\end{tabular}}
\end{center}
\caption{Results on the ICDAR15 test set under different model configurations and training strategies. AP numbers are reported. ``PD'', ``Mask'' and ``Inc-Res'' stand for partially labeled data, RoI masking and Inception-ResNet respectively. ``Det-baseline'' refers to the first step (training the detector using fully labeled data) of the ``two-step'' training process.}
\label{table:onevstwo}
\end{table}

\noindent \textbf{Partially Labeled Data:} The use of partially labeled data provides significant improvements in end-to-end performance across all configurations of our model (row 4 \vs row 6, or row 3 \vs row 5). Interestingly, \emph{it also improves the detector without training the detection branch directly} (Section~\ref{sec:loss}). Once again, this suggests that we can improve the feature extractor by receiving training signal through recognition branch.

\noindent \textbf{RoI Masking:} In Table ~\ref{table:onevstwo}, we show the effectiveness of RoI masking (row 3 \vs row 4, or row 5 \vs row 6). Higher end-to-end AP scores are consistently achieved in the presence of RoI masking (\eg +3.3\% AP with the ResNet-50 backbone when using partially labeled data).  This demonstrates that the recognizer benefits from RoI masking. The improvement is more significant for a lighter weight backbone with smaller receptive field. For detection performance, we observe mixed results: a marginal improvement for the Inception-ResNet backbone, and some degradation when using ResNet-50.

\noindent \textbf{Training Strategy:} Row 2 and row 6
of Table ~\ref{table:onevstwo} compare the effect of single-step and two-step training strategies as described in Section \ref{sec:details}. In single-step training, we jointly optimize the detector and recognizer together using both fully and partially labeled data. In two-step training, we first train the detection-only baseline, and then add a recognizer into joint training. We find that single-step training consistently out-performs two-step training in both detection and end-to-end evaluations. Single step training is far simpler, and makes it easier to apply automatic hyperparameter tuning and neural architecture search, which we will study in future work.




\subsection{Speed}
For images from the ICDAR15 dataset (with resolution $1280\times 720$), the end-to-end inference time is 210 ms for the ResNet-50 backbone and 330 ms for the Inception-ResNet backbone (on a Tesla V100 GPU). If we only run the detection branch, the corresponding inference time are 180 ms and 270 ms respectively. Thus, for scene text images, the computational overhead of the recognition branch is quite small. Sharing the same CNN feature extractor makes end-to-end model more computationally efficient than two-step methods.

\section{Conclusion}
In this paper, we present an end-to-end trainable network that can simultaneously detect and recognize text in arbitrary shape. The use of Mask R-CNN, attention decoder and a simple yet effective RoI masking step leads to a flexible and high performance model. We also show that end-to-end training can benefit from partially machine annotated data. On the ICDAR15 and Total-Text benchmarks, our method significantly surpasses previous methods by a large margin while being reasonably efficient.

{\small
\bibliographystyle{ieee_fullname}
\bibliography{iccv19}
}

\end{document}